\documentclass[11pt]{article}
\usepackage{coling2020}
\usepackage{times}
\usepackage{url}
\usepackage{latexsym}
\usepackage{xcolor}
 
 \newcommand{\mg}[1]{{\textcolor{olive}{#1}}}

\usepackage[normalem]{ulem} 
\usepackage{multirow}
\usepackage{soul}

\newcommand{\invisible}[1]{{}}
\usepackage{xcolor,pifont}
\newcommand*\colourcheck[1]{%
  \expandafter\newcommand\csname #1check\endcsname{\textcolor{#1}{\ding{52}}}%
}
\colourcheck{black}

\colingfinalcopy % Uncomment this line for the final submission

\title{Breeding Gender-aware Direct Speech Translation Systems }

\author{Marco Gaido\textsuperscript{1,2 $\dagger$}, Beatrice Savoldi\textsuperscript{2 $\dagger$}, Luisa Bentivogli\textsuperscript{1}, Matteo Negri\textsuperscript{1}, Marco Turchi\textsuperscript{1} \\
  \textsuperscript{1}Fondazione Bruno Kessler, Trento, Italy \\
  \textsuperscript{2}University of Trento, Italy \\
  {\tt \{mgaido,bentivo,negri,turchi\}@fbk.eu,beatrice.savoldi@unitn.it} \\}

\date{}

\begin{document}
\maketitle
\begin{abstract}
%It should be no longer than 200 words.

\blfootnote{\textsuperscript{$\dagger$}The authors contributed equally.}

In automatic speech translation (ST), traditional \textit{cascade} approaches involving separate transcription and translation steps are
%\bs{\erase{gradually}}
giving ground to increasingly competitive and 
%%%potentially%%%
more robust \textit{direct} solutions.  In particular, by translating speech audio data without intermediate transcription, direct
%approaches to ST
ST models are able to 
%preserve and leverage
leverage
and preserve essential  information present in the  input (e.g. speaker's vocal characteristics)
%traits) 
that is otherwise lost in the 
cascade framework. Although such ability proved to be useful for gender translation, direct ST is nonetheless affected by gender bias just like its cascade counterpart,  as well as machine translation
and numerous other natural language processing applications. Moreover, direct ST systems that exclusively rely on vocal biometric features as a gender cue can be unsuitable
%%% or even
and potentially harmful for certain users.
Going beyond speech signals, in this paper we compare different approaches to inform  direct ST models about the speaker's gender and test their ability to
handle gender translation from English into Italian and French. To this aim, we manually annotated large datasets with speakers' gender information and used them for  experiments reflecting different possible real-world scenarios. Our results show that gender-aware direct ST solutions can significantly outperform strong -- but gender-unaware -- direct ST models. In particular, the translation of gender-marked words can increase up to 30 points in accuracy while preserving overall translation quality.

\end{abstract}

\section{Introduction}
\label{sec:intro}

\blfootnote{
    \hspace{-0.65cm}  % space normally used by the marker
    This work is licensed under a Creative Commons 
    Attribution 4.0 International License.
    License details:
    \url{http://creativecommons.org/licenses/by/4.0/}.
}

Language use is intrinsically social and situated as it varies across groups and even individuals \cite{bamman-etal-2014-distributed}.
 As a result, the language data that are collected to build the
 %large
 corpora on which 
natural language processing 
%(NLP)
models are trained are often far from being homogeneous and rarely offer a fair representation of different demographic groups and their linguistic behaviours \cite{bender-friedman-2018-data}. Consequently,  as  predictive  models learn  from the data distribution they have seen,
%\mg{questo però non è vero: i modelli non tendono a overgeneralizzare perché imparano dai dati che vedono....}
%they tend to overgeneralize and 
they tend to favor the demographic group most represented in their training data \cite{hovy-spruit-2016-social,shah-etal-2020-predictive}.
 
This brings serious social consequences as well, since the people who are more likely to be underrepresented within datasets are those whose representation is often less accounted for within our society.
%\cite{bender-friedman-2018-data} \bs{questa reference non con questa frase}.
%\mg{so che mi odierete ma io lo dico lo stesso: stiamo scrivendo un paper su gender bias for ST. Dopo mezza pagina di introduzione questo è il primo riferimento a uno dei topic del paper... A me sta cosa perplime, ma sono un ingegnere e non capisco una sega...}
A case in point regards the gender data gap.\footnote{For a comprehensive overview on such societal issue see \cite{criado2019invisible}.}
%data gap
%gap.\footnote{\cite{criado2019invisible}} 
In fact, studies on speech taggers 
\cite{hovy-sogaard-2015-tagging}
and speech recognition \cite{tatman-2017-gender} showed that the underrepresentation of female speakers in the training data
leads to significantly lower accuracy in modeling 
%such a 
that demographic group. 

The problem of gender-related differences has also been inspected within automatic translation, both from text \cite{vanmassenhove-etal-2018-getting}
and from audio \cite{bentivogli-etal-2020-gender}.
These studies -- focused on the translation of spoken language -- revealed a systemic gender bias whenever
systems are
required to overtly and formally express speaker's gender in the target languages, while translating from languages that do not convey such information.
Indeed, languages with
%a 
grammatical
%system of
gender, such as French and Italian, display a complex morphosyntactic and semantic system of gender agreement \cite{Hockett:58,Corbett:91}, relying on feminine/masculine markings
that reflect
%reflecting
speakers' gender on numerous parts of speech whenever they are talking about themselves 
%(e.g. 
%(\mn{what we call ``speaker-dependent'' gender expressions like ``\textit{stata''/``\textit{stato''} in}}
(e.g. En: \textit{I've never \textbf{been} there} -- It: \textit{Non ci sono mai {\textbf{stata/stato}}}). Differently, English is a natural gender language \cite{Hellinger} that mostly conveys gender via its pronoun system, 
%\mg{cite?}
but only for third-person pronouns (\textit{he/she}), thus to refer to an entity other than the speaker. 
As the example shows,  in absence of contextual information  (e.g \textit{As a woman, I have never been there})
%pronouns indicating the gender identity of the speaker and the referred entities) 
correctly translating gender
%can become
can be prohibitive. This is the case of traditional  text-to-text  machine translation (MT) and of the so-called \textit{cascade} approaches to speech-to-text translation (ST),
which involve
%involving
separate transcription and translation 
steps \cite{StentifordSteer88,Waibel1991b}.
%Although \textit{direct} approaches \cite{berard_2016,weiss2017sequence},
%which translate without intermediate transcriptions,
%are 
%partially capable of extracting useful information from the 
%input
%(e.g. by inferring speaker's gender from 
%his/her
%vocal traits), the general problem persists:
Instead, \textit{direct} approaches \cite{berard_2016,weiss2017sequence} 
translate without intermediate transcriptions.
Although this makes them partially capable of extracting useful information from the input
(e.g. by inferring speaker's gender from his/her vocal characteristics),
%traits
the general problem persists:
since
%feminine markerd-words
female speakers (and associated feminine marked words)
%\mg{\erase{associated}}
are less frequent within the
%data used to train such models,
training corpora, automatic translation tends towards a masculine default. 

Following \cite{crawford2017trouble},  this attested systemic bias  can directly affect the users of such  technology by diminishing their gender identity  or further exacerbating  existing social inequalities and access to opportunities for women.  Systematic gender representation problems -- although unintended -- can affect users’ self-esteem \cite{Bourguignon:15}, especially when the linguistic bias is shaped as a perpetuation of stereotypical gender roles and associations \cite{Levesque2011}. Additionally, as the system does not perform equally well across gender groups, such tools may not be suitable for women, excluding them from benefiting from new technological resources. 
%}

To 
%this day,
date, few attempts have been made 
%toward the development of speaker gender-aware translation models,
towards developing gender-aware translation models, and surprisingly, almost exclusively 
within the MT community
\cite{vanmassenhove-etal-2018-getting,Elaraby2018GenderAS,moryossef-etal-2019-filling}. 
%Instead, the 
%Bentivogli et al. \shortcite{bentivogli-etal-2020-gender}, the
The only work on gender bias in ST \cite{bentivogli-etal-2020-gender} proved that direct ST has an advantage when it comes to 
%speaker-dependent gender translation,
%\mn{translating speaker-dependent gender-marked expressions (e.g. \textit{``I am a \textbf{teacher}''} uttered by a woman),} 
speaker-dependent gender translation 
%(e.g. translating  \textit{``I am a \textbf{teacher}''} uttered by a woman into Italian),
%(as in \textit{``I am a \textbf{teacher}''} uttered by a woman),}
(as in \textit{I've never \textbf{been} there} uttered by a woman),
since it can leverage acoustic  properties from the audio input (e.g. speaker's fundamental frequency). %While such finding attests the superiority of direct ST for this specific \bs {phenomenon}, the authors leave some questions unattended. Namely, their work does not consider that 
However, relying on 
%typical pitch differences as perceptual markers of speakers’ gender
perceptual markers of speakers’ gender is not the best solution for all kinds of users (e.g. transgenders, 
%individual with vocal impairment).
children, vocally-impaired people).
Moreover, although their conclusions remark that direct ST is nonetheless affected by gender bias, no attempt has yet been made to try and enhance its gender translation 
capability.
Following these observations, and considering that ST applications have entered widespread societal use, we believe that more effort should be put into further investigating and controlling gender translation in direct ST, in particular when the gender of the speaker is known in advance.

%ORIGINALE 
% Towards this objective, we annotated a large ST dataset  with speakers' gender information and explored different techniques to exploit such information in direct ST.

Towards this objective, we annotated MuST-C~\cite{mustc,MuST-Cjournal} - the largest freely available multilingual corpus for ST - with speakers' gender information and explored different techniques to exploit such information in direct ST.
The proposed techniques are compared, both in terms of overall translation quality as well as accuracy in the translation of gender-marked words, against a ``pure'' model that solely relies on the speakers' vocal characteristics for gender disambiguation.

In light of the above, our contributions are: 

\noindent
\textbf{(1)} the manual annotation of the TED talks contained in MuST-C with speakers' gender information, based on the personal pronouns found in their TED profile. The resource is released under a CC BY NC ND 4.0 International license, and is freely downloadable at \url{https://ict.fbk.eu/must-speakers/};

% \textbf{(1)} the \bs{thorough} annotation of \bs{speakers' gender
% \footnote{\bs{This resource is released under a CC BY NC ND 4.0 International license, and is freely downloadable at \url{https://ict.fbk.eu/must-speakers/}}}
% based on their personal pronouns} \mg{in the TED profile}
% for the TED talks contained in 
% MuST-C~\cite{mustc,MuST-Cjournal}, the largest freely available multilingual corpus for ST;

\noindent
\textbf{(2)} the first comprehensive exploration of different approaches to 
mitigate gender bias
%\mg{help and control the gender of speaker's related terms} ==> LB: non lo capisco, era piu' semplice prima
%\mg{Luisa: è più semplice, ma è corretto? è la riduzione del gender bias il grosso punto?}
in direct ST, depending on the potential users, the available resources and the architectural implications of each choice.

Experiments carried out on English-Italian and English-French
%en-it and en-fr
show that, on both language directions,
%Our results show that,  on both language directions,
our gender-aware systems
%, each with its own advantages and disadvantages,
significantly outperform 
%strong ``pure'' (i.e. gender-unaware) 
``pure'' 
%(i.e. gender-unaware) 
ST models
%, with overall improvements 
in the translation of gender-marked words (up to 30 points in accuracy) while preserving overall translation quality.
%In particular, our specialized systems achieve huge improvements in terms of gender treatment while preserving overall translation quality
Moreover, our 
best
%\mg{specialized} 
systems learn to produce feminine/masculine gender forms regardless of the perceptual features 
%they are receiving 
received from
%\mg{in} 
the audio signal, offering a solution for
%\mg{\erase{the above-mentioned}}
cases where relying on speakers' vocal characteristics is detrimental to a proper gender translation.
%%%of their \mg{preferred linguistic gender.}
%are not indicative of their gender identity. 

\section{Background}
\label{sec:rel-work}
Besides the abundant work carried out for English monolingual NLP tasks \cite{sun-etal-2019-mitigating},
%\bs{sono tanti citarne almeno qualcuno},, 
a consistent amount of studies have now inspected how MT is affected by the problem of gender bias.
Most of them, however, do not focus on 
% the \mg{\erase{generation}} 
speaker-dependent gender agreement. Rather, a number of studies \cite{stanovsky-etal-2019-evaluating,escude-font-costa-jussa-2019-equalizing,saunders-byrne-2020-reducing} evaluate whether MT is able to associate prononimal coreference with an occupational noun 
%(in an English source sentence) 
to produce the correct masculine/feminine forms in the target gender-inflected languages 
% (En: I’ve known \textit{her} for a long time, \textbf{my friend}
% %works as an 
% is a \textbf{cook.} Es: \textit{La} conozco desde hace mucho tiempo, \textbf{mi amiga} 
% %trabaja como 
% es \textbf{cocinera}). 
(En: \textit{I’ve known \underline{her}  for a long time, \textbf{my friend}
%works as an 
is a \textbf{cook.}} Es: \textit{\underline{La} conozco desde hace mucho tiempo, \textbf{mi amiga} 
%trabaja como 
es \textbf{cocinera}}).

Notably, few approaches have been employed to make neural MT systems speaker-aware by controlling 
%the gender translation hypothesis.\lb{non capisco} 
gender realization in their output. \newcite{Elaraby2018GenderAS} enrich their data with a set of gender-agreement rules so to force the system to account for them in the prediction step. In~\cite{vanmassenhove-etal-2018-getting},
the MT system is augmented at training time by prepending a gender token (female or male) to each source segment. Similarly, \newcite{moryossef-etal-2019-filling}
%\mg{\erase{adopt a black-box approach and}}
artificially inject a short phrase (e.g. \textit{she said}) at inference time, which acts as a gender domain label for the entire sentence. 
These approaches are implemented and tested
on natural spoken language that,
%with respect 
compared to written language,  is more likely to contain references to the speaker and, consequently, 
speaker-dependent gender-marked words.

In the light of above, the correct translation of gender is a particularly relevant task for ST systems, as they are precisely developed to translate oral, conversational language. Nonetheless, to our knowledge only one work has investigated gender bias in ST \cite{bentivogli-etal-2020-gender}. 
Focusing on the proper handling of gender 
%phenomena in translation, 
phenomena,
%\cite{bentivogli-etal-2020-gender} 
the authors
take stock of the situation by comparing cascade and direct 
%(end-to-end) 
architectures on MuST-SHE, a multilingual benchmark derived from the TED-based MuST-C corpus
%~\cite{mustc,MuST-Cjournal} 
and specifically designed to evaluate gender translation and bias in ST. Their conclusions remark that, although traditional cascade systems still outperform  direct solutions, the latter are able to  exploit audio information  
for a better treatment of speaker-dependent gender phenomena.

These findings open a line  of focused research on speaker-aware ST that is worth exploring more thoroughly, also in light of the fact that the %gap between the two technologies
performance gap between  cascade and direct approaches
has further  reduced~\cite{ansari-etal-2020-findings}.
On one side, rather than comparing the two paradigms, this progress now motivates exploring 
% %all the possible ways to boost direct ST performance.
all the possible ways to boost direct ST performance towards the translation of gender-marked 
%expressions referring to the speaker. 
expressions. On the other side,  since the direct systems tested in \cite{bentivogli-etal-2020-gender} rely on ``pure'' models built to verify an hypothesis (i.e. that translating audio signals without intermediate representations makes a difference 
%when translating gender-marked expressions),
in handling gender), the real potential of direct ST technology with respect to this problem is still unknown.  Moreover, as their ``pure'' models solely rely on the speaker's fundamental frequency, various instances in which such perceptual marker is not indicative of the speaker's gender 
%identity 
remain out of the picture.

\section{Annotation of MuST-C with Speakers' Gender Information}
\label{sec:data}

Although current research on gender-aware ST can  count on the MuST-SHE benchmark~\cite{bentivogli-etal-2020-gender} for fine-grained evaluations, gender-annotated training data are not yet available.  
So far, this has limited
%%%This limits 
the scope of research to application scenarios in which speakers' gender
%identity 
%has to be
is inferred from the input audio. These scenarios are not representative of the full range of possible usages of ST and are also potentially problematic, since gendered forms expected in translation do not necessarily align with speaker's vocal characteristics.

% \mg{is} inferred from the input audio, which are not representative of the full range of possible usages of ST \mg{and is potentially harmful, as gendered forms expected in translation do not necessarily align with speaker's vocal characteristics (e leverei la frase dopo)}. 
%\bs{is inferred from the input audio, which are not representative of the full range of possible usages of ST. Moreover, such scenarios erroneously presuppose that the set of gendered forms expected in translation can be attributed from -- and somewhat necessarily align with -- speakers' vocal characteristics.}
%%%In light of this our goal of maximizing the accuracy of gender treatment in operating conditions where speakers' gender identity is known advocates for  building large training corpora explicitly annotated with such information.
In the light of the above, building large training corpora explicitly annotated with gender information becomes crucial.
%
%\bs{In the light of the above, our goal of maximizing the accuracy of gender treatment in operating conditions where speaker's \mg{\st{most likely}} preferred gendered linguistic forms are known advocates for building large training corpora explicitly annotated with such information.}
To this aim, rather than building a new resource from scratch, we opted for adding an annotation layer to MuST-C,
%~\cite{mustc,MuST-Cjournal},
which  has been chosen over other existing corpora \cite{europarlst} for the following reasons:
\textit{i)} it is currently the largest freely available multilingual corpus for ST, \textit{ii)} being based on TED talks it is the most compatible one with MuST-SHE, 
%%% and
 \textit{iii)}  TED speakers' personal information is publicly available and retrievable on the TED official website.\footnote{Available at \url{https://www.ted.com/speakers/}}
 %, \bs{\textit{iv)} we consider TED as an authoritative and validated source.} \mg{non vedo il motivo di questa aggiunta: il Parlamento Europeo non lo è?}

Following the MuST-C talk IDs, 
%for the two language pairs of our interest (en-it/en-fr), 
we have been able to  \textit{i)} automatically retrieve the speakers' name, \textit{ii}) find their associated TED official page, and \textit{iii})  manually label %%%their gender based on 
the personal pronouns used in their descriptions.
Though time-consuming, such manual retrieval of information is preferable  
%%%has been preferred 
to automatic speaker gender identification 
%\mg{qua va bene parlare di gender identification?}
%in order to conduct sound experiments under optimal conditions.
for the following reasons. 
%%%two reasons.
First, since automatic methods based on fundamental frequency are not equally accurate across demographic groups 
%as when differentiating the both typically high women/children's pitch -- \cite{Levitan:2016}}
(e.g. women and children are hard to distinguish as their pitch is typically high \cite{Levitan:2016}), 
%\cite{Levitan:2016}%
%accurate (e.g. in distinguishing children/women's pitch (typically high) \cite{Levitan:2016})
%accurate, --
manual assignment prevents from incorporating 
%noise (i.e. 
gender misclassifications in our training data.
%Additionally, 
Second, biological essentialist frameworks that categorize gender based on acoustic cues \cite{zimmantransgender} are especially problematic for transgender individuals, whose gender identity is not aligned with the sex they have been assigned at birth based on designated anatomical/biological criteria \cite{stryker2008transgender}.

Differently, following the guidelines in \cite{larson-2017gender}, we do not want to run the risk of making assumptions about speakers' gender identity
%exclusively 
%based on their biometric features 
and introducing additional bias within an environment that has been specifically designed to inspect gender bias.
% Rather,  by looking at the personal pronouns used by the speakers to describe themselves, our manual assignment better accounts for the gender linguistic forms by which the speakers accept to be referred to \cite{GLAAD}.
%
%Rather, 
By looking at the personal pronouns used by the speakers to describe themselves, our manual assignment instead is meant to account for the gender linguistic forms by which the speakers accept to be referred to in English \cite{GLAAD}, and would want their translations to conform to.
%In fact, speakers' personal pronouns ultimately is the most relevant information for identifying the most likely grammatical gender form that the speakers would want their translations to conform to. 
%
%Note that, w
We stress that gendered linguistic expressions do not directly map to speakers' self-determined gender identity \cite{cao-daume-iii-2020-toward}. 
 We therefore make explicit that throughout the paper, when talking about speakers' gender, we refer to their accepted linguistic expression of gender rather than their gender identity.
\begin{table}
\centering
\small
\begin{tabular}{|l| c c | c c | c c|} 
 \hline
 & Talks M & Talks F & Hours M & Hours F & Segments M & Segments F \\
 \hline
 \textbf{en-it} & 1,569 & 725 & 316 & 136 & 178,841 & 71,877 \\ 
 \textbf{en-fr} & 1,569 & 725 & 327 & 151 & 189,742 & 81,527 \\
 \hline
\end{tabular}
\caption{
%MuST-C overall number of Talks, total numbers of segments, segments with Female and Male speaker for both \mg{en-it} and \mg{en-fr}.
Statistics for MuST-C data with gender annotation. The number of segments and hours varies over the two language pairs due to the different pre-processing of MuST-C data.}
  \label{tab:MuST-C_annotatated_stats}
\end{table}

Focusing on the two language pairs of our interest, 2,294 different speakers described via \textit{he/she} pronouns\footnote{
%We
%don't
%\mg{do not} imply the necessary conflation of chromosomal/biological sex (assigned at birth) and self-ascribed female/male gender identity.
It is important to point out that some individuals  do not neatly fall into
%adhere to 
the female/male binary 
%gender 
(gender fluid, non-binary) or may even not experience gender at all (a-gender) \cite{richards2016non,Schilt,GLAAD}, possibly preferring the use of singular \textit{they} or other neopronouns.
Within MuST-C,  speakers with \textit{they} pronoun have been encountered, but MuST-C human-reference translations do not exhibit linguistic gender-neutralization strategies, which are difficult to fully implement in languages with grammatical gender \cite{lessinger2020challenges}. Note that, because of such inconsistency and the very limited number of cases, these instances were not used 
for training.
Our experiments therefore focus on binary 
%gender identities and 
linguistic forms.
By design, some sparse talks with multiple speakers of different genders were also excluded. 
Detailed information about all MuST-C speakers and corresponding talks can be found in the resource release at \url{ict.fbk.eu/must-speakers}.} are represented in both en-it and en-fr.
Their male/female\footnote{Some authors distinguish female/male for sex and woman/man
for gender (among others \cite{larson-2017gender}). For the sake of simplicity, in our study we use female/male to respectively indicate those speakers whose personal pronouns are \textit{she/he}.}
% Following \cite{kaufmann2014masculine,cao-daume-iii-2020-toward}, we do not require %nor foreground 
% this distinction in our study and also rely on female/male for gender when referring to the speakers.\lb{attenzione, qui va cambiato
distribution is unbalanced, 
%for both language pairs
as shown in Table \ref{tab:MuST-C_annotatated_stats}, which presents the number of talks, as well as the total number of segments and the corresponding hours of speech.

\section{ST Systems}
\label{sec:systems}

For our experiments, we built three types of direct systems. One is the
\textit{base} system, a state-of-the-art model that does not leverage any external information about speaker's gender ($\S$\ref{subsec:base}). The others are  two gender-aware systems that exploit speakers' gender information in different ways:  \textit{multi-gender} ($\S$\ref{subsec:prepending}) and \textit{specialized} ($\S$\ref{subsec:finetuning}).
%
%Our model is
All the models share the same architecture, a Transformer \cite{vaswani2017attention} adapted to ST.
%\mg{\erase{The memory usage required by a Transformer depends quadratically
%on the length of the input; since audio sequences are much longer than
%their textual form, this architecture requires solutions to reduce the input length to be used in an ST scenario.
%For this reason,}}
%\mg{In particular,}
The encoder processes the input Mel-filter-bank sequences with two 2D convolutional layers with stride 2,
returning a sequence 
%four times shorter.
that is four times shorter than the original input.
%to fit network's memory.
The vectors of this sequence are projected by a linear transformation
into the dimensional space used in the following encoder Transformer layers and are summed with sinusoidal positional embeddings.
The attentions in the encoder layers are biased toward elements close on the time dimension with a logarithmic distance penalty \cite{diGangi2019enhancing}.
The decoder architecture, instead, is not modified.

\subsection{Base ST Model}
\label{subsec:base}

We are interested in evaluating and improving gender translation on strong ST models that can be used in real-world contexts. As such, our base,
%%% PRECEDENTE %%% Our base
gender-unaware model is trained
%using the techniques which showed to be effective
with the goal of achieving state-of-the-art performance on the ST task. 
%In particular, we rely on the data augmentation
To this aim, we rely on  data augmentation  and knowledge transfer techniques that were shown to 
%be essential for the quality of the resulting model in
yield competitive models at
%  the IWSLT-2020 evaluation campaign \cite{ansari-etal-2020-findings,potapczyk-przybysz-2020-srpols,gaido-etal-2020-end}.
the IWSLT-2020 evaluation campaign \cite{ansari-etal-2020-findings,potapczyk-przybysz-2020-srpols,gaido-etal-2020-end}.
%
%The three data augmentation methods we used are: SpecAugment~\cite{Park_2019}, time stretch~\cite{nguyen2019improving}, and synthetic data generation \cite{jia2018leveraging}. We transfer knowledge both from ASR and MT. 
In particular, we use
three data augmentation methods -- SpecAugment~\cite{Park_2019}, 
time stretch~\cite{nguyen2019improving}, and synthetic data generation \cite{jia2018leveraging} -- and we  transfer knowledge both from ASR
and MT 
%(with knowledge distillation~\cite{hinton2015distilling}).
%and knowledge transfer from ASR and MT --
%performed through component initialization and knowledge distillation~\cite{hinton2015distilling}.
through component initialization and knowledge distillation~\cite{hinton2015distilling}.

%Indeed, the 
The ST model's encoder is initialized with the encoder of an 
%acoustic 
English 
ASR model \cite{bansal-etal-2019-pre} with a lower number of encoder layers (the missing layers are initialized randomly, as well as the decoder). 
%The acoustic English ASR 
This ASR model is trained on Librispeech \cite{librispeech}, Mozilla Common Voice,\footnote{\url{https://voice.mozilla.org/}} How2 \cite{sanabria18how2}, TEDLIUM-v3 \cite{Hernandez_2018}, and the utterance-transcript pairs of the ST corpora --
Europarl-ST \cite{europarlst} and
%MuST-C \cite{mustc,MuST-Cjournal}.
MuST-C.
These datasets are either gender unbalanced or do not provide speaker's gender information 
%for their samples
% apart from Librispeech, which is balanced in terms of}  \bs{female/male speaker, but here the speakers are just a book's narrator, thus first-person sentences do not 
% %really
% refer to the speakers themselves.}
apart from Librispeech, which is balanced in terms of  female/male speakers~\cite{garnerin-etal-2020-gender}. However, since 
%here the 
these speakers are just  book  narrators,  first-person sentences do not really refer to the speakers themselves.

%Moreover, we distill knowledge 
Knowledge distillation (KD) is performed
from a \textit{teacher} MT model by optimizing the cross entropy
between the distribution produced by the teacher and 
%\st{that produced}
by the \textit{student} ST model being trained \cite{liu2019endtoend}.
For both en-it and en-fr, the MT model is trained on the OPUS datasets  \cite{opus}.

The ST model is trained in three consecutive steps.
%In the first phase, the input data are the ASR audio samples and the targets are the translations of the corresponding transcripts generated by an MT model.
In the first step,
%the input data consist of ASR audio samples paired with the translations of the corresponding transcripts generated by an MT model.
we use the synthetic data obtained by pairing ASR audio samples  with the automatic 
%(hence potentially noisy) 
translations of the corresponding transcripts.
%This training distills knowledge from a \textit{teacher} MT model by optimizing the cross entropy
%between the distribution produced by the teacher and that produced by the \textit{student} ST model being trained
%\cite{liu2019endtoend}.
%
%In the second  phase, the model is trained on the ST corpora.
In the second step, the model is trained on the ST corpora.
In these first two steps, we use the KD loss function.
Finally, in the third step, the model is fine-tuned on the same ST corpora using label-smoothed cross entropy~\cite{szegedy2016rethinking}.
SpecAugment and time stretch are used in all steps.

\subsection{Multi-gender Systems}
\label{subsec:prepending}

The idea of ``multi-gender'' models, i.e. models informed about the speaker's gender with a tag prepended to the source sentence, was introduced by \newcite{vanmassenhove-etal-2018-getting} and \newcite{Elaraby2018GenderAS}. This approach was inspired by one-to-many multilingual neural MT systems \cite{johnson-etal-2017-googles}, in which a single model 
is trained to translate from a source  into many target languages by means of a \textit{target-forcing} mechanism.
With this mechanism - here adapted for ``\textit{gender-forcing}'' - ST multi-gender systems are fed not only with the input audio,
but also with a tag (\textit{token}) representing the speaker's gender.
This \textit{token} is converted into a vector through learnable embeddings.
%There are several options, though, to provide the model with the additional gender information. In this paper, we consider those  
This approach has two main potential advantages: \textit{i)} a single model supports both male and female speakers (which makes it particularly appealing for real-world application scenarios), and \textit{ii)} each gender direction can benefit from the data available for the other, potentially learning to produce words that would have never been seen otherwise (\textit{transfer learning}).
%Among
Regarding the several options to supply the model with the additional gender information, we do not follow the approach of \newcite{vanmassenhove-etal-2018-getting} and~\newcite{Elaraby2018GenderAS}, since it is dedicated to MT. Instead,
we consider those that obtained the best results in multilingual direct ST~\cite{digangi2019onetomany,hirofumi2019_multilingual}, namely:

\noindent\textbf{Decoder prepending.} The gender token replaces the \textit{\textless\textbackslash s\textgreater} (\textit{EOS}, end-of-sentence) that is added in front of the generated tokens in the decoder input.

\noindent\textbf{Decoder merge.} The gender embedding is added to all the word embeddings representing the generated tokens in the decoder input.

\noindent\textbf{Encoder merge.} The gender embedding is added to the Mel-filter-bank sequence 
%that represents the source speech and is the input of the encoder.
representing the source speech  given as input to the encoder.

%\mt{Quale dei metodi sopra appartiene a Di Gangi e quale a Inaguma? Io ci metterei accando la citazione.}

In all cases, multi-gender models' weights are initialized with those of the \textit{Base} models.
The only randomly-initialized 
%learnable
parameters are those of the gender embeddings.
% During training, 
% % the mini-batches are all composed of the same number of samples for each gender. 
% all the the mini-batches contain the same number of samples for each gender. 
% In case one gender has more training data than the other,
% the under-represented gender pairs are over-sampled to have a balanced training set.

%Recent work  proposes different strategies to apply the target forcing mechanism to multilingual end-to-end speech translation~\cite{digangi2019one}. 
 
% i) concat, which prepends a language embedding to the input sequence, and ii) merge, which sums the embedding to all the elements in the sequence. 

%We replicated these approaches in order to assess if they lead to differences with respect to the treatment of gender phenomena. 

\subsection{Gender-specialized Systems}
\label{subsec:finetuning}

In this approach, two different gender-specific models are created.
Each model is initialized with the \textit{Base} model's weights and then fine-tuned
only on
%audio segments \mt{Sembra che usiamo solo l'audio: ``audio-translation pairs''}
samples of the corresponding speaker's gender.
This solution has the drawback of a higher maintenance burden than the multi-gender one, as it requires the training and management of two 
%different
separate models.
Moreover, no transfer learning is possible: although each model is initialized with the base model trained on all the data and the low learning rate used in the fine-tuning prevents catastrophic forgetting \cite{mccloskey:catastrophic},
data scarcity conditions for a specific gender are likely to lead to 
lower performance on that direction.

\subsection{Gender-balanced Validation Set}
\label{subsec:new-dev-set}

To train our gender-aware models, we do not rely on the standard MuST-C validation set as it reflects the same  gender-imbalanced distribution found in the training data. We therefore created a new specifically designed validation set composed of 20 talks.  Unlike the standard MuST-C validation 
set, it contains a balanced number of female/male speakers, thus avoiding to reward models' potentially biased behaviour. This new resource is released under a CC BY NC ND 4.0 International license, and is freely downloadable at \url{https://ict.fbk.eu/must-c-gender-dev-set/}.\footnote{To ease future research on gender bias in ST for the three language pairs represented in MuST-SHE (en-it, en-fr, en-es), the validation set is also available for en-es.}

\section{Experimental Setting}

\subsection{Experiments}

As described in $\S$\ref{subsec:base}, our ST models adopt knowledge transfer techniques that showed to significantly improve ST performance. In particular, knowledge distillation (KD) is especially relevant as it allows the ST model to learn and exploit the wealth of training data available for MT,
%as much as possible from the MT training data,
which otherwise would not be accessible.
%\mn{MT training data}.
%Despite the important gains in terms of translation quality, in this work 
Hence, since we are also interested in assessing the effect of KD
%distilling knowledge from the MT model 
on the ability of the resulting ST systems to deal with gender, we compare: \textit{i)} the teacher MT models, \textit{ii)} the intermediate ST models trained on 
%knowledge distillation
KD, and \textit{iii)} the final ST models obtained with fine-tuning without 
%knowledge distillation
KD.
%Since we are interested in assessing the contribution of knowledge coming from MT data on the ability of ST systems to deal with gender translation, we add to our experiments the output of \textit{i)}  the MT system used for knowledge distillation, and \textit{ii)} the intermediate training steps of the \textit{base} model that rely specifically on knowledge distillation.\\

The final ST 
%model was
models are used to initialize both multi-gender ($\S$\ref{subsec:prepending}) and gender-specialized models ($\S$\ref{subsec:finetuning}), which
%were
are then fine-tuned
%with gender-labeled data.
on the MuST-C gender-labeled dataset.
% As seen in $\S$\ref{sec:data}, this dataset shows a quite skewed male/female speaker distribution, amounting to around 30\%/70\% in terms of samples. Thus, we tested both approaches in two different data conditions: \textit{i)} a balanced condition, where we used all the female data available together with a random subset of the male data (referred as \textsc{*-Bal}); and \textit{ii)} an unbalanced condition where all the MuST-C data available are exploited (referred as \textsc{*-All}).
Since, as seen in $\S$\ref{sec:data}, this dataset shows a quite skewed male/female speaker distribution (approximately 70\%/30\%),  we test both approaches in two different data conditions: \textit{i)} 
%a balanced condition 
balanced (\textsc{*-Bal}), where we use all the female data available together with a random subset of the male data, and \textit{ii)} 
%an unbalanced condition 
unbalanced (\textsc{*-All}) where all the MuST-C data available are exploited.
It must be noted that there are differences between the two approaches on the usage of data. In the specialized approach, since we have two separate systems, the one which is fine-tuned with talks by female speakers remains the same in both data conditions.
Differently, in the multi-gender approach, which is trained on both genders together, all the training mini-batches 
%\lb{must?}
contain the same number of samples for each gender. Thus, when all MuST-C data are used, the female gender pairs -- which are underrepresented -- are over-sampled.

\subsection{Evaluation Method}
\label{subsec:method}

For our experiments, we rely on MuST-SHE~\cite{bentivogli-etal-2020-gender}, a gender-sensitive, multilingual benchmark for MT and ST consisting of (\textit{audio, transcript, translation)} aligned triplets. By design, each segment in the corpus requires the translation of at least one English gender-neutral word into the corresponding masculine/feminine target word(s) to convey a referent's gender.
%
%It is compiled with segments that require the translation of at least one English gender-neutral word into the corresponding masculine or feminine target word(s).
%
%In order to
%
%%%To evaluate our gender-aware ST models on speaker-dependent gender phenomena%%%, 
With the intent to evaluate our gender-aware ST models on speaker-dependent gender phenomena, we focus on a portion of MuST-SHE  containing, for each language pair, $\sim$600 segments where gender agreement
%agreement 
only depends on the speaker's gender.\footnote{In ~\cite{bentivogli-etal-2020-gender} this portion is referred to as ``Category 1''.} 
Segments are balanced with respect to 
%the amount of 
female/male speakers and  masculine/feminine marked words, which are explicitly  annotated in the corpus.

An important feature of MuST-SHE is that, for each reference translation, an almost identical ``wrong'' reference is created by swapping each annotated gender-marked word into its opposite gender 
(e.g. \textit{I have been} uttered by a woman is translated into the correct Italian reference \textit{Sono stat\textbf{a}}, and into the wrong reference \textit{Sono stat\textbf{o}}).
%\bs{Both words with wrong and correct gender-marking are annotated within the corpus}.
The idea behind gender-swapping is that 
%scores' difference
the difference between the scores
computed against
%between 
the ``correct'' and the ``wrong'' reference sets captures the system's ability to handle gender translation.
However, relying on  
%this single score
these scores
does not allow to distinguish between those cases where 
%system's output contains 
the system ``fails'' by producing
a word 
%that is 
different from 
the one
%annotated word
%gender-marked target 
 present in the references  (e.g. \textit{andat\textbf{*}} in place of \textit{stat\textbf{*}})
%present in MuST-SHE 
and failures specifically due to the wrong realization of gender (e.g. \textit{stat\textbf{o}} in place of \textit{stat\textbf{a}}).

Thus, while following the same principles
%and criteria 
as~\newcite{bentivogli-etal-2020-gender},
in our experiments we rely on a more informative evaluation.
%For each system, we calculate its \textbf{term coverage} in MuST-SHE, i.e. 
First, we calculate the \textbf{term coverage}
%, i.e.
as
the proportion of gender-marked words annotated in MuST-SHE that are actually generated by the system, on which the accuracy of gender realization is therefore \textit{measurable}. 
Then, %among these 
%gender-marked 
%generated
%words,
we define \textbf{gender accuracy} as the 
%ratio
proportion
of correct gender realizations among the words on which it is \textit{measurable}.
%correctly generated by the system \mg{(}disregarding the correctness of the gender\mg{)}.
%
Our evaluation method has several advantages.
On one side,
\textit{term coverage} unveils the precise amount of words on which systems' gender realization is measurable.
On the other, \textit{gender accuracy}  directly informs about systems' performance on gender translation and related gender bias: 
%gender accuracy scores go from 0 to 100 and do not have negative values, however
% it ranges from 0 to 100 and 
scores below 50\% indicate that the system produces the wrong gender more often than the correct one, thus signalling a particularly strong bias.
%Gender accuracy scores have
Gender accuracy has the further advantage of informing about the margins for improvement of the systems.

\section{Results}
\label{sec:results}

\subsection{Overall Results} 

Table~\ref{tab:bleu} presents overall results in terms of BLEU scores on the MuST-SHE test set. Despite the well-known  differences in performance between en-it and en-fr,
%\bs\{the well-known difference in absolute values} between en-it and en-fr 
both language directions show the same trend. 

First, the \textsc{MT} systems used by the ST models for KD achieve by far the highest performance. This is expected since the ST task is more  complex and MT models are trained on larger amounts of data.
However, all
%Nevertheless, it is worth remarking that all 
our ST results are  competitive compared to those published for the two target languages. In particular, on the 
%\bs{official} 
MuST-C test set, the scores of our ST \textsc{Base} models are  27.7 (en-it) and 40.3 (en-fr), respectively 0.3 and 4.8 BLEU points above the best \textit{cascade} results reported in \cite{bentivogli-etal-2020-gender}.

%% ENIT: 21.5 direct --- 27.4 cascade
%% ENFR 31.0 direct --- 35.5 cascade

%%%%%%%%%%%%%%%%%%%%%%%%%%%%%%%%%%%%%%%%%%%%%%%%%%%%%%%%%%%%%%
%%%%%%%%%% BLEU ON MUST-SHE  + ACCURACY ON CAT 1 %%%%%%%%%%%%%
%%%%%%%%%%%%%%%%%%%%%%%%%%%%%%%%%%%%%%%%%%%%%%%%%%%%%%%%%%%%%%

\begin{table}[b]
\parbox{.40\linewidth}{
\centering
\small
\begin{tabular}{|l|c||c|}
\cline{2-3} 
 \multicolumn{1}{c|}{} & \multicolumn{1}{c||}{\textbf{En-It}} & \multicolumn{1}{c|}{\textbf{En-Fr}}  \\ \cline{2-3}
  
 \multicolumn{1}{c|}{} & Bleu &  Bleu    \\ \hline

MT for KD           & 33.59 & 39.61\\
\hline
Base-KD-only	    & 23.58 & 31.97\\
Base               & 27.51 & 34.25\\
\hline
Multi-DecPrep-Bal          & 26.36 & 33.54\\
Multi-DecPrep-All         & 26.17 & 34.13\\
Multi-EncMerge-Bal           & 26.47 & 33.29\\
Multi-EncMerge-All          & 26.39 & 33.07\\
Multi-DecMerge-Bal           & 21.99 & 27.06\\
Multi-DecMerge-All          & 22.12 & 27.74\\
\hline
Specialized-Bal    & 27.43 & 34.32\\
Specialized-All   & 27.79 & 34.61\\
\hline 
\end{tabular}
\caption{BLEU scores on MuST-SHE.}
\label{tab:bleu}
}
\hfill
\parbox{.50\linewidth}{
\centering
\small
\begin{tabular}{|l|cc||cc|}
\cline{2-5} 
 \multicolumn{1}{c|}{} & \multicolumn{2}{c||}{\textbf{En-It}} & \multicolumn{2}{c|}{\textbf{En-Fr}}  \\ \cline{2-5}
  
 \multicolumn{1}{c|}{} & Cover. & Acc.  & Cover. & Acc.    \\ \cline{1-5} 

MT for KD &  63.83 & 	51.45 & 63.10 & 	52.08 \\ 
\hline

Base-KD-only &  56.05 & 	51.76 & 		59.17 & 	53.12  \\ 

Base &  56.17 & 	56.26 & 		62.02 & 	56.24 \\ 
\hline

Multi-DecPrep-Bal  &  56.91 &	64.86 & 60.95 & 69.34 \\ 

Multi-DecPrep-All & 56.54 & 	66.81  & 61.31 & 	70.29 \\

Multi-EncMerge-Bal &  57.04 & 	62.55  & 		60.60 & 	62.67 \\

Multi-EncMerge-All &  57.65 & 	60.39  & 	62.38 & 	61.83 \\

Multi-DecMerge-Bal  &  49.88 & 	59.41	  & 54.52 & 	64.63 \\

Multi-DecMerge-All  &   50.74 & 	60.58	  & 56.31 & 	65.96\\ 
\hline

Specialized-Bal  &   57.90 & 	86.35	  & 61.79 & 	86.13 \\

Specialized-All  &   58.02 & 	87.02  & 	62.38 & 	86.45 \\ 
\hline

\end{tabular}
\caption{Term coverage and gender accuracy scores.}
\label{tab:accuracy-cat1}
}
\end{table}

%%%%%%%%%%%%%%%%%%%%%%%

Moving on to ST systems, we attest that the models after the first two training steps
%of the model
based on KD (\textsc{Base-KD-only}, see \ref{subsec:base})
%, which rely on knowledge distillation,
have a lower translation quality than the 
%final
\textsc{Base} models, showing that the third training step is crucial to boost overall performance. 
%\mt{Io metterei un rimando a dove si spiegano i tre step di training... altrimenti qui si rischia che se ne sia gia' dimenticato}
%
In general,
%\mn{However,}
  except for the \textsc{Multi-DecMerge} system (whose performance is significantly lower),
%\mn{in general} 
we do not observe statistically significant differences between the 
%\textit{base}
\textsc{Base}
models and their gender-aware extensions 
%(\textit{Multi-gender} and  \textit{Specialized})
(\textsc{Multi-*} and \textsc{Specialized-*}), which also perform on par when fine-tuned with 
%different
varying amounts of annotated data 
%(50 vs. 100).
(balanced vs all).
%\mt{In general, i Multi sono sempre sotto le performance del base di piu' di un punto BLEU.}
%
% CANCELLATO DA MN
%\bs{The only exception is given by the \textsc{Multi-DecMerge} system, whose performance is significantly lower.}
%\mg{For this reason our recommendation is to prefer the other two solutions.}
%\mt{Scusate se rompo le scatole, ma se i nostri sistemi sono molto piu' bravi del Base a generare le parole presenti nella reference, perche' abbiamo una riduzione seppur minima di BLEU? Secondo me, quando avete finito di infamarmi, potrebbe avere senso discutere questa cosa... secondo me la motivazione e' che forzare il gender comporta dei cambiamenti che non sono solo nelle parole che marcano il gender, ma anche in altre, questo puo' causare delle traduzioni che si allontanano dalla reference.}

%
%
%To
% As BLEU is not informative about the specific phenomenon 
% %of gender realization 
% due to the very small percentage of words expressing speaker's gender-marking
% ($< 3\%$, 810-840 over $\sim$30,000 words in MuST-SHE). 

Due to the very small percentage of speaker-dependent gender-marked words 
%expressing speaker's gender 
in MuST-SHE
($< 3\%$, 810-840 over $\sim$30,000 words), systems' ability to translate gender is not reflected by BLUE scores. 
Now, we delve deeper into our 
%analysis of gender translation.
more informative evaluation (as per $\S$\ref{subsec:method}) and turn to the term coverage and gender accuracy values presented in Table~\ref{tab:accuracy-cat1}.
%,
%\mg{As BLEU is not informative about this phenomenon due to the very small percentage of words reflecting it
%($< 3\%$, 810-840 over $\sim$30,000 words),}
%we \bs{must}
%now
%turn to \mg{the} term coverage and gender accuracy scores
%\mg{reported in Table~\ref{tab:accuracy-cat1}.}
%\mg{Indeed, BLEU 
%does not show the differences in
%\bs{is not informative about} the ability of the system to account for the speaker's gender,
%as the number of words reflecting it in the test set (810-840) is a very small percentage ($< 3\%$,  MuST-SHE contains about 30.000 words)}.
%which are exclusively focused on gender-marked words. 
%
%Table~\ref{tab:accuracy-cat1} shows the same trend for both en-it and en-fr.
%
%
The overall results assessed with BLEU are confirmed by \textbf{term coverage} scores
for both en-it and en-fr:
the MT systems 
%\bs{\erase{are able to}}
generate the highest number of 
%gender-marked
annotated words present in MuST-SHE
%\mg{\erase{, disregarding the correctness of gender realization}} 
(63.83\% on en-it and 63.10\% on en-fr),
while
%the different ST approaches do not show relevant differences
%\mg{there are not relevant differences among ST models}
we do not observe large
 differences among the ST models (between 56.17\% and 58.02\% for en-it and 60.60\% and 62.38\% for en-fr).
Instead, looking at \textbf{gender accuracy}, we immediately unveil that overall performance is not an indicator of the systems' ability to translate gender. In fact, the best performing MT systems show the lowest gender accuracy (51.45\% for en-it and 52.08\% for en-fr): intrinsically constrained by the lack of access to audio information, they produce the wrong target gender in half of the cases. 
%in half of the cases \bs{\erase{the}} systems produce
%translate with
%the wrong gender.
Such deficiency
%behaviour
is directly reflected in the \textsc{Base-KD-only} models, which
%exploiting as much as possible the MT knowledge it
%\mg{learn to mimic the MT behaviour;}
are strongly influenced by the MT behaviour;
%\bs{\erase{and also inherit
%its
%their biases}}
thus, although effective for overall quality,
KD is detrimental to gender translation.
% \mg{\erase{it is necessary to be aware that,}} while
%Knowledge Distillation
%\mg{KD} is an effective method to improve the overall quality of direct models,
%this method
%\mg{it} is detrimental to gender translation.
%
%ACC TRA LE LINGUE E' UGUALE NONOSTANTE IL FRANCESE ABBIA PERFORMANCE MOLTO MIGLIORI. Quindi vediamo anche cross-lingually che tradurre meglio NON significa tradurre meglio il genere.
%
By undergoing %third training step,
the third 
%fine-tuning
training step
without KD,
the
%final
\textsc{Base}
%model is
models are in fact able to improve
%its performance
on gender translation\mg{,} but 
%without relevant 
with limited
gains.
Differently, the models fed with the speaker's gender information display a noticeable increase in gender translation, with \textsc{Specialized-*} models outperforming the \textsc{Multi-*} ones by 16--20 points and the \textsc{Base} ones by 30 points.

Among the 
%\bs{\erase{different}} 
multi-gender architectures, %approach demonstrate that the
our results show that \textsc{Multi-DecPrep} has an edge on the other two models, both 
%as
in
overall and gender translation 
%\mg{\erase{performance:}}
performance: for the sake of simplicity, from now on we thus present only that model.
%\mn{hence discuss} only that model.
%\mn{hence proceed with}  only that model.
As a single-model architecture, multi-gender
%While the single multi-gender architecture
would be a more
%efficient
functional solution
%with respect to
than multiple specialized models,
%the fact that this model is trained
but -- being trained on both female and male speakers' utterances --
%\mg{\erase{, which are also balanced,}}
%makes
%it noticeably less strong in terms of ability to translate gender than a single system trained only with data of same-gender speakers.
it is noticeably weaker than multiple specialized models (trained on gender-specific data) at predicting gender.
%Since the results obtained up to know \mt{now?} on the Multi-gender
%
%
%\mt{Secondo me da qualche parte, va detto che rispetto al modello Base noi abbiamo una condizione aggiuntiva, ovvero noi dobbiamo sempre sapere il gender dello speaker, sia per il Multi che per lo specialized.}
%
With regard to the different amounts of gender-annotated data
used to train our
gender-aware models, we cannot see any appreciable
%difference
variation in term coverage and gender accuracy between the two settings.
%both 
%in terms of
%\bs{for term} coverage and gender accuracy.
Further insights on this aspect
%on this topic 
are presented in the 
%following fine-grained evaluation.}
next section.
%To have additional clues on this issue, we rely on the fine-grained evaluation presented below.

%\mg{\erase{In order to be able to understand the extent to which the different ST systems exhibit a gender bias, we now carry out a more fine-grained evaluation by separating feminine and masculine word forms.}}
%, which are determined directly by the speaker's gender.

%Since the results obtained up to know \mt{now?} on the Multi-gender approach demonstrate that the \textsc{Multi-DecPrep} model has an edge on the other two models, both as overall  and gender translation performance, for the sake of simplicity in the following analysis we present only that model. \mg{però allora questo andrebbe sottolineato prima anche, cosa che non viene fatta... oppure lo sottlineiamo pirma e lo leviamo da qua}

%%%%%%%%%%%%%%%%%%%%%%%%%%%%%%%%

\begin{table}[b]
\centering
\small
\begin{tabular}{|l|cc|cc| c |cc|cc|}
\cline{2-5} \cline{7-10}
 \multicolumn{1}{c|}{} & \multicolumn{4}{c|}{\textbf{En-it}} &  & \multicolumn{4}{c|}{\textbf{En-Fr}} \\ \cline{2-5} \cline{7-10}
 
 \multicolumn{1}{c|}{} & \multicolumn{2}{c|}{\textbf{Feminine}} & \multicolumn{2}{c|}{\textbf{Masculine}} & 
  & \multicolumn{2}{c|}{\textbf{Feminine}} & \multicolumn{2}{c|}{\textbf{Masculine}} \\ \cline{2-5} \cline{7-10} 
  
 \multicolumn{1}{c|}{} & Cover. & Acc.  & Cover. & Acc.  
 &  & 
 Cover. & Acc.   & Cover. & Acc.    \\ \cline{1-5} \cline{7-10}

MT for KD &  66.25 & 	16.23 & 		61.46 & 	88.49  
&  & 
63.76 & 	16.24 & 		62.41 & 	89.58   \\ \cline{1-5} \cline{7-10}

Base-KD-only &  58.75 & 	20.85 & 		53.41 & 	84.93 
&  & 
58.59 & 	26.91 & 		59.76 & 	79.44  \\ 

Base &  58.75 & 	33.62 & 		53.66 & 	80.45  
&  & 
60.47 & 	32.30 & 		63.61 & 	79.55 \\ \cline{1-5} \cline{7-10}

Multi-DecPrep-Bal  &  60.00 & 	68.75 & 		53.90 & 	60.63
&  & 
61.41 & 	68.58 & 		60.48 & 	70.12  \\ 

Multi-DecPrep-All &  58.00 & 	69.83 & 		55.12 & 	63.72  
&  & 
61.88 & 	65.78 & 		60.72 & 	75.00  \\

%Multi-EncMerge-50 &  59.75 & 	66.95 & 		54.39 & 	57.85 
%&  & 
%60.47 & 	65.37 & 		60.72 & 	59.92 \\

%Multi-EncMerge-100 &  61.00 & 	59.43 & 		54.39 & 	61.43  
%&  & 
%62.59 & 	66.17 & 		62.17 & 	57.36 \\

%Multi-DecMerge-50  &  48.00 & 	49.48 & 		51.71 & 	68.40  
%&  & 
%55.76 & 	63.71 & 		53.25 & 	65.61  \\

%Multi-DecMerge-100  &  46.50 & 	48.92 & 		54.88 & 	70.22  
%&  & 
%56.71 & 	64.32 & 		55.90 & 	67.67  \\ 
\cline{1-5} \cline{7-10}

Specialized-Bal &  62.00 & 	79.84 & 		53.90 & 	93.67
&  & 
62.59 & 	79.32 & 		60.96 & 	93.28 \\

Specialized-All  &  62.00 & 	79.84 & 		54.15 & 	95.05 
&  & 
62.59 & 	79.32 & 		62.17 & 	93.80  \\ \cline{1-5} \cline{7-10}

\end{tabular}
\caption{Coverage and accuracy scores divided by feminine and masculine word forms.}
  \label{tab:accuracy-cat1-MF}
\end{table}

\subsection{Cross-gender Analysis}
Table \ref{tab:accuracy-cat1-MF} shows separate term coverage and gender accuracy scores
%split according to feminine and masculine word forms
%separated 
for target feminine and masculine forms.
%In this way, we are able 
This allows us to highlight the models'
%ability to represent
translation ability for each gender form and
%the corresponding bias.
conduct cross-gender comparisons to detect potential bias.
Also in this analysis, results are consistent across language pairs.
We 
%can 
assess that both the \textsc{MT} model and its strongly connected \textsc{Base-KD-only} present a very strong bias since they almost always produce masculine forms:
%. Indeed, gender
accuracy is always much lower than 50\% on the feminine set (up to 20.85\% for en-it and  26.91\% for en-fr) and
%it is 
very high on the masculine set (up to 88.49\% for en-it and  89.58\% for en-fr). 
After fine-tuning without 
%knowledge distillation,
KD, the \textsc{Base} ST 
%model improves
models improve
%on the female representation, 
feminine forms realization,
%; yet still, it remains
but they still remain
%\mn{; yet still, they remain}
far 
%even
from 50\%.
% Moreover, 
% %its
% \mn{their}
% ability to handle gender translation is poorer compared with the model in \newcite{bentivogli-etal-2020-gender}, whose overall translation quality is however much lower. Once again, this confirms that higher overall performance do not directly imply a better speaker' gender treatment; rather, we asses that KD actually hinders it. 
%
The comparison with the direct model in  \cite{bentivogli-etal-2020-gender} shows that, despite the much higher overall translation quality, our \textsc{Base} models are affected by a stronger bias. This further confirms the detrimental effect of KD on gender translation and that higher overall quality does not directly imply a better speaker's gender treatment.

All gender-aware models significantly reduce bias with respect to the  \textsc{Base} systems. This is particularly evident in the feminine set, where accuracy scores far above 50\% 
%showing
indicate their ability to correctly represent female speakers. 
%discriminate the 
%\mg{\erase{correct}}.
In particular, the \textsc{Specialized} models achieve the best results 
%overall 
on both feminine and masculine sets (over 79\% and 93\% respectively).
%In particular, the \textsc{Specialized} models achieve the best results overall (over 79\% and 90\% respectively) and improve over the \textsc{Base} systems on both the feminine and masculine sets.
% The better performance on the masculine set can \bs{at least be partially explained} \bs{accounting for the strong bias of the \textsc{Base} model, from which the two gender-specialized models are then derived. }
%
%%\bs{The higher performance on the masculine set can be explained in the light of the strong bias towards masculine form of the \textsc{Base} model, from which the other two gender-specialized models are derived.}
%
The higher performance on the masculine set can be  explained considering that the two gender-specialized models derive from the \textsc{Base} model, which is strongly biased towards masculine forms. 
Interestingly, 
\textsc{Multi-DecPrep} shows similar feminine/masculine accuracy scores. 
This is possibly due to the random 
%(hence unbiased) 
initialization of the gender tokens' embeddings: as a result, the initial model hidden representations and predictions are
%random as well.
perturbed in an unbiased way. An unbiased starting condition combined with
balanced data
leads to a fairer, similar behaviour across genders, although
the final models have a lower accuracy than the \textsc{Specialized} ones.

Finally, we notice that results obtained by training our models with balanced (\textsc{*-Bal}) and unbalanced (\textsc{*-All}) datasets are similar. Indeed, the masculine gender accuracy slightly improves by adding more male data, while there is not a clear trend on the feminine accuracy: we can conclude that oversampling the data is functional inasmuch it keeps the performance on the feminine set stable.

\subsection{Analysing Conflicts between Vocal %Traits 
Characteristics and Gender Tags}
%\subsection{\mn{Analysing Voice/Gender Tags Conflicts}}

So far, we worked under the assumption that the speaker's vocal characteristics
%traits
match with those typically associated to the gender category she/he identifies with.
In this section, we explore
%ability
systems' capacity 
to produce
%of producing 
%a translation that is
translations that are
coherent with the speaker's gender in a 
%analyze the scenarios 
%consider a 
scenario
%who need an interpreter.
in which this assumption does not hold:
this is the case of some transgenders, children and people with vocal impairment.
%\noindent\textbf{Gender swap analysis.} 
%As a final analysis, we explore
%%ability
%systems' capacity 
%to produce
%%of producing 
%%a translation that is
%translations that are
%coherent with the speaker's gender, but under 
%this
%a
%new scenario:
%\mg{the speaker's gender differs from the one typically associated to the vocal traits.}
%the speaker's gender does not match the typical vocal traits associated with that gender.
%%\mn{scenario in which}
%the speaker's gender 
%%identity
%does not match the typical vocal traits 
%%perceptual 
%%markers 
%associated with that gender.
%\mg{the speaker's gender differs from the one typically associated to the vocal traits.}
%with female/male voices. 
%This is the case of transgenders, children and 
%interpreters of people with vocal impairment and children, too.
%people with vocal impairment who need an interpreter.
%\mn{with 
%typical 
%female/male vocal traits (as in the case of transgenders
%, interpreters of people with vocal impairment 
%and children).}
%\mg{capacity} of systems to produce a 
%translation which reflects the gender identity of the speaker in those scenarios where it does not correspond to the typical perceptual markers associated with
%female/male
%\mg{the} voice
%(e.g. in case of transgender persons, children, or interpreters of individuals with vocal impairment). 
However, we are hindered by the almost absent representation of such %scenario/
users within MuST-C.
%\mg{MuST-C.} 
As such, we design a counterfactual experiment where we associate the opposite gender tag to each actual female/male speaker and inspect models' behaviour when receiving 
%dissociated 
conflicting
information between the gender tag and the properties of the acoustic signal. 
%. Accordingly, we inspect how tag information 
%\mg{and we inspect models' behavior when the tag information is} dissociated from the expected acoustic signal.
%interact with each other.
%
%
%
%
% On one side, this can be considered as \mn{an} indirect assessment of 
% %robustness of the systems 
% \mn{systems' robustness}
% to possible errors in the identification of the speaker's gender. 
% % On the other side,  we aim to shed some light on alternative solutions for direct ST whenever relying on acoustic features 
% % %
% % %from the acoustic signal 
% % does not represent the best solutions for speaker gender translation.
% On the other,  
% %we aim 
% \mn{this allows us}
% to shed some light on alternative solutions for direct ST whenever relying on acoustic features does not represent the best solution for speaker gender translation.
%
This can also be considered as an indirect assessment of 
%robustness of the systems 
systems' robustness  
%either to actual  conflicts or 
to possible errors 
%in the identification of the speaker's gender in application scenarios where 
in application scenarios where speakers' gender is assigned automatically.

Table~\ref{tab:accuracy-cat1-swapped} presents the results for this experiment. In the \textit{M-audio/F-transl} set, 
%the system was 
systems were
fed with a male voice and a female tag and the expected translation is in the feminine form, while in the  \textit{F-audio/M-transl} set we have the opposite. 
% As shown in Table~\ref{tab:accuracy-cat1-swapped}, in the \mg{\textit{M-audio/F-transl}} set \mn{(i.e. ``masculine-like'' voice for a female speaker)} the system was fed with a male voice and a female tag, and the expected translation is in the feminine form, while in the  \mg{\textit{F-audio/M-transl}} set we have the opposite. 
%
%
%giving to each system the opposite gender of the speaker and considering as correct reference the one 
%referred as ``Wrong'' in MuST-SHE, ie. the one which corresponds 
%corresponding to the provided gender token and not to the speaker's voice.
%
As we can see, in both sets the multi-gender model has a drastic drop in accuracy with respect to the 
%standard evaluation (see Table~\ref{tab:accuracy-cat1-MF})
results shown in Table \ref{tab:accuracy-cat1-MF},
%in both Table~\ref{tab:accuracy-cat1-MF}
%\mg{\erase{Feminine and Masculine}}
 with scores below 50\% for en-it. 
This behaviour indicates that this model relies on both the gender token and the audio features, which in this scenario are
%in contrast. 
conflicting. Thus, the multi-gender model could be more robust to possible errors in automatic recognition of the speaker's gender, but it is not usable in scenarios in which the 
%gender expressed by the speaker's  voice has to be be ignored.
vocal characteristics
%traits
have to be be ignored.
On the contrary, the specialized systems show a high accuracy on both %\mg{\erase{the feminine and masculine}} 
sets. In particular, on
%the masculine set
\textit{F-audio/M-transl}
the performance is 
%the same as in the standard evaluation.
in line with the results of Table \ref{tab:accuracy-cat1-MF}.
\invisible{\footnote{The lower performance on %the feminine set
\mg{\textit{M-audio/F-transl}} can be ascribed to the fact that this model is trained on half the number of female speakers with respect to the model specialized on male speakers.}}
% This 
% %demonstrates
% \mn{indicates}
% that the model relies only on the provided gender information to produce the translation, without considering the voice, being therefore suitable to translate in situations in which 
% %we want
% \mn{one wants}
% to control the speaker's gender identity.
 This indicates that, independently from speakers' vocal characteristics, the model relies only on the provided gender information,
 %
 %As such, it goes beyond speech signals and is able to  produce the desired gender forms. 
%
 being therefore suitable for situations in which  one wants to control the gendered forms in the output and override the potentially misleading speech signals. 

%being therefore suitable to translate   in situations in which  one wants to control the gender forms in the output. 

% %we want
% \mn{one wants}
% to control the speaker's gender identity
%able to produce the desired gender forms in translation. 
%suitable to
%translate in situations in which one wants to control the speaker's gender identity.
% \mg{able to translate according to the selected preferred linguistic gender indicated by the speaker.}
%
%\bs{(non so sia il caso di dire che ne ho contati 5...magari erano di piu' ma le annotazioni delle prime talk non li hanno riportati. Idem dire che io ho trovato 17 talk con transgender speaker su tutto TED,   magari ce ne sono altre ma non sono indicate esplicitamente }

\begin{table}[h]
\centering
\small
\begin{tabular}{|l|cc|cc| c |cc|cc|}
\cline{2-5} \cline{7-10}
 \multicolumn{1}{c|}{} & \multicolumn{4}{c|}{\textbf{En-it}} &  & \multicolumn{4}{c|}{\textbf{En-Fr}} \\ \cline{2-5} \cline{7-10}
 
% \multicolumn{1}{c|}{} & \multicolumn{2}{c|}{\textbf{Feminine}} & \multicolumn{2}{c|}{\textbf{Masculine}} & 
%  & \multicolumn{2}{c|}{\textbf{Feminine}} & \multicolumn{2}{c|}{\textbf{Masculine}} \\ 
  %\cline{2-5} \cline{7-10} 
  
  \multicolumn{1}{c|}{} & \multicolumn{2}{c|}{\textbf{M-audio/F-transl.}} & \multicolumn{2}{c|}{\textbf{F-audio/M-transl.}} & 
  & \multicolumn{2}{c|}{\textbf{M-audio/F-transl.}} & \multicolumn{2}{c|}{\textbf{F-audio/M-transl.}} \\ \cline{2-5} \cline{7-10} 
  
 \multicolumn{1}{c|}{} & Cover. & Acc.  & Cover. & Acc.  
 &  & 
 Cover. & Acc.   & Cover. & Acc.    \\ \cline{1-5} \cline{7-10}

\cline{1-5} \cline{7-10}
Multi-DecPrep-All  & 		54.88 &	45.78 &  60.25 &	38.17
&  & 
61.93 &	45.14 & 61.18 &	55.77 		  \\
\cline{1-5} \cline{7-10}

Specialized-All  &  54.39 &	64.57 & 60.75 &	94.24 
&  & 
62.17 &	59.69 & 61.41 &	94.25  \\ \cline{1-5} \cline{7-10}

\end{tabular}
\caption{Coverage and accuracy scores when the correct translation is expected in a gender form opposite to the speaker's gender but in accordance with the gender tag fed to the system.}
  \label{tab:accuracy-cat1-swapped}
\end{table}

%M: parla una femmina, tag maschio, traduzione attesa maschile

%F: parla un maschio, tag femmina, traduzione attesa femminile.

%multi-normale	1F 58.00\%	69.83\%		1M 55.12\%	63.72\%

%multi-swapped	1F 60.25\%	38.17\%		1M 54.88\%	45.78\%

%specialized-normale	1F 62.00\%	79.84\%		1M 54.15\%	95.05\%

%specialized-swapped	1F 60.75\%	94.24\%		1M 54.39\%	64.57\%				

\section{Manual Analysis}
\label{sec:analysis}

We complement our automatic evaluation with a manual inspection on the output of three models:  
% \textit{Base}, \textit{Multi-DecPrep-100} \mn{(\textsc{Multi-100})} and \textit{Specialized-100} \mn{(\textsc{Spec-100})}. 
\textsc{Base}, \textsc{Multi-DecPrep-All} (\textsc{Multi}), and \textsc{Specialized-All} (\textsc{Spec}). For each model, we 
%\mg{\erase{manually}}
analyzed the translation of 100 common segments across en-it/en-fr, 
%thus allowing 
which allow for cross-lingual comparisons. 
%For comparative purposes, we inspect en-it/en-fr translated segments from the MuST-SHE common subset.  
%

% \begin{table*}[h!]
\begin{table*}[t!]
\setlength{\tabcolsep}{5pt}
 \centering
\small
\begin{tabular}{lllll}
\hline
 %\textbf{Form}
%&  &  \\ [1.mm]\\

(a) F & IT & \textsc{src} & In one, I was \textbf{the classic Asian student}...&  \\
& & \textsc{ref} & In uno ero \textbf{la classica studentessa asiatica}...&  \\
& & \textsc{Base}  &  In una, ero \textbf{il classico studente asiatico}...& \ding{55} \ding{55} \ding{55} \ding{55} \\
& & \textsc{Multi} \& \textsc{Spec} & Innanzitutto, ero \textbf{la classica studentessa asiatica}... & \blackcheck\blackcheck\blackcheck\blackcheck\\ %[1.5mm] 
\hline
% (b) Fem. & \textsc{src} & As a \textbf{researcher}, \textbf{professor} and \textbf{new parent}...   \\
(b) F & FR & \textsc{src} & As a \textbf{researcher}, a \textbf{professor}... & \\
%and new parent...
& & \textsc{ref} & En tant que \textbf{chercheuse}, \textbf{professeure}... & \\
% & \textsc{Base}$_{FR}$ &  En tant que \textbf{chercheur}, \textbf{professeur} et \textbf{nouveau parent}...& \textcolor{red}{\ding{55}} \textcolor{red}{\ding{55}} \textcolor{red}{\ding{55}} \\
& & \textsc{Base} &  En tant que \textbf{chercheur}, \textbf{professeur}... &
%et nouveau parent...
\ding{55}  \ding{55} \\
% & \textsc{Multi-100} \& \textsc{Spec-100}$_{FR}$ & En tant que \textbf{chercheuse}, \textbf{professeur} et  \textbf{nouveau parent}...& \greencheck  \textcolor{red}{\ding{55}}  \textcolor{red}{\ding{55}} \\ [1.5mm] 
& & \textsc{Multi} \& \textsc{Spec} & En tant que \textbf{chercheuse}, \textbf{professeur}...
%et  nouveau parent...
& \blackcheck   \ding{55} \\ %[1.5mm] 

%(c) M & \textsc{src} & And just as the \underline{woman} who wanted to know me as an \textbf{adult}... & \multirow{3}{1.7cm}{adulto} &\\
%& \textsc{Base} \& \textsc{Multi} &  E proprio come una donna che voleva conoscermi da \textbf{adulta}...& & \textcolor{red}{\ding{55}}  \\ 
%& \textsc{Spec} &  E proprio come una donna che voleva conoscermi da \textbf{adulto}...  & & \greencheck  \\[1.5mm]
\hline

(c) M & IT  & \textsc{src} & ...the \underline{woman} who wanted to know me as an \textbf{adult}... & \\
 & & \textsc{ref} & ...la \underline{donna} che voleva vedere come fossi diventato da \textbf{adulto}... & \\
& & \textsc{Base} \& \textsc{Multi} &  ... una donna che voleva conoscermi da \textbf{adulta}...& \ding{55}  \\ 
& & \textsc{Spec} &  ... una donna che voleva conoscermi da \textbf{adulto}...  & \blackcheck  \\ %[1.5mm]
\hline

(d) F & FR & \textsc{src} & When I was a \textbf{kid}... &\\
& & \textsc{ref} & Quand j'étais \textbf{petite}... & \\
& & \textsc{Base} \& \textsc{Multi} \& \textsc{Spec}  &  Quand j' ai été \textbf{tuée}... & \textbf{?}  \\ %[1.5mm] 
\hline

(e) F & IT & \textsc{src} & ...while downhill skiing \textbf{paralyzed} me. &\\
& & \textsc{ref} & ...quando un terrificante incidente sciistico mi ha lasciato \textbf{paralizzata}. &\\
& & \textsc{Base} \& \textsc{Multi} \& \textsc{Spec}  &  ... quando mi \textbf{paralizzò}. & \textbf{?}  \\ %[1.5mm] 
\hline

(f) F & IT & \textsc{src} & I was \textbf{elected}... &\\
& & \textsc{ref} & Sono stata \textbf{elett\textbf{a}}...  &\\
& & \textsc{Base} &  Fui \textbf{elett\textbf{o}}... & \ding{55}  \\
& & \textsc{Multi} \& \textsc{Spec} &  Fui \textbf{selezionata}...  &  \textbf{?}  \\ 
\hline

\end{tabular}
\caption{
%Examples of \textit{measurable} 
%feminine/masculine gender-marked words comparing
Examples of 
feminine (F) and masculine (M) gender-marked words translated by
\textsc{Base}, \textsc{Multi-DecPrep-All} (\textsc{Multi}) and \textsc{Specialized-All} (\textsc{Spec}) %translation 
%outputs
on en-it and en-fr.}
\label{tab:Examples}
\end{table*}

%
%
%
%To gain a better insight into the effect of our modeling strategy on the specific phenomenon of gender translation, we first take into account those instances were the production of gender-marked words was evaluated (see Table \ref{tab:Examples}).
% \mn{better understand}
%\mn{To gain insights into their behaviour regarding the specific phenomenon of gender translation,} we

We first take into account those instances where systems' accuracy in the production of gender-marked words was \textit{\textit{measurable}},
%as in the examples provided
as in (a), (b), (c)
in Table \ref{tab:Examples}.
%(see Table \ref{tab:Examples}).
%
A first observation,
%which is
consistent across languages and models, is that a 
%translated 
controlling noun (\textit{student}) and its modifiers (\textit{the, classic, Asian}) always concord in gender in the systems' output. As per (a), this agreement is respected for both correct
%-feminine 
(\textsc{Multi}, \textsc{Spec}) and wrong
%-masculine 
%3gender 
gender realizations (\textsc{Base}).
Differently, (b) shows that, whenever two words are not related by any morphosyntactic dependency, %\mg{\erase{then}} within the same sentence 
some words may be correctly translated (\textit{chercheuse} -- \textsc{Multi}, \textsc{Spec}),
%next to words with incorrect masculine forms.
and some others not (\textit{professeur}). 
%\mg{even though they are next to words with incorrect form.}
Such dynamic seems to attest that,
%in spite of receiving
%specific gender data/tokes applying to the whole sentence,
although the systems 
%receive 
are fed with
sentence-level gender tags,
gender predictions are still skewed at the level of the single word.
%\mg{\erase{Except when confronted with related words, where agreement rules are inferred by the models and reflected  in the consistent distribution of the same gender forms.}}
%
% Overall (a),(b),c), we clearly attest 
Overall, (a), (b) and (c) clearly attest the progressively improved performance from \textsc{Base} to \textsc{Multi} and \textsc{Spec}.  In particular, in (c), 
%specialized models are
\textsc{Spec} is
able to pick the required masculine form in spite of a contextual hint about a second female referent (\textit{woman}), thus overcoming what is a difficult prediction even for \textsc{Multi}.
%, too.

%Moreover, we inspect those cases where the production of gender marking
We also inspect those cases where 
%the production of gender-marked words
systems' accuracy on gender production
was not \textit{measurable} to cast some light on the reasons for a
%narrow
limited \textit{term coverge}. 
%Accordingly, we
We found that, while there are some generally
%inadequate 
wrong 
translations -- (d) --
%of near homophones  
%(e.g. \textit{I was a kid } translated into \textit{j'ai \'{e}t\'{e} tu\'{e}} -- \textit{I was killed}),
%(e.g. \textit{I was a kid } translated as if the source were \textit{I was killed}), 
such instances only amount to 1/3 of 
%such non-measurable sample.
the cases. 
%More often, 
In the remaining 2/3,
the output is fluent and reflects the source utterance meaning but it simply
%doesn't
does not match the exact annotated word in the reference.
%, thus their evaluation. 
We found that ST translations often offer alternative constructions that
%don't
do not require an overt gender-inflection
%(e.g. \textit{paralized}: REF--\textit{paralizz\textbf{ata}} vs. ST-- \textit{paralizz{\`o}}),
-- (e) --
or rely on appropriate gender-marked synonyms of the word in the reference 
%(e.g. \textit{elected}: REF--\textit{scelt\textbf{o}} vs. ST--\textit{elett\textbf{o}}).
-- (f).
We can hence conclude that many gender translations 
%excluded from \textit{gender accuracy} confirm
that do not contribute to
\textit{gender accuracy} 
confirm
%conceal
%We can hence conclude that many gender translations excluded from \textit{term coverage} conceal
an improved ability of the enriched models in gender translation.

\section{Conclusion}
%In this work, we have accepted and 
We rose to the challenge posed by \newcite{bentivogli-etal-2020-gender} to further explore gender translation 
%within \erase{the paradigm of} 
in
direct 
ST.
%speech translation.
%\erase{approaches to} ST. 
Going beyond direct 
%system's
systems' attested ability
%of leveraging 
to leverage
speaker's 
vocal characteristics
%traits
from the audio input, we developed gender-aware 
models 
%in
suitable for
operating conditions where speaker's gender 
%identity
%\mg{preferred linguistic gender} 
is known. To this aim, we annotated 
the large  MuST-C dataset
%~\cite{mustc,MuST-Cjournal} 
with speaker's gender 
information, and used the new annotations to experiment with
different architectural solutions: ``multi-gender'' and ``specialized''.
%Without affecting overall translation quality, our
Our results on two language pairs (en-it and en-fr) show that  breeding speaker's gender-aware ST improves the correct realization of 
%masculine/feminine 
gender. In particular, our specialized systems outperform the gender-unaware ST models
by 30 points in gender accuracy without affecting overall translation quality.

\section*{Acknowledgements}
We would like to thank the anonymous reviewers and the COLING'2020 Ethics Advisory Group for their insightful comments.
This work is part of the ``End-to-end Spoken Language Translation in Rich Data Conditions'' project,\footnote{\url{https://ict.fbk.eu/units-hlt-mt-e2eslt/}} which is financially supported by an Amazon AWS ML Grant.

\bibliographystyle{coling}
\bibliography{coling2020}

\end{document}